# A Large Multi-Target Dataset of Common Bengali Handwritten Graphemes

[1,2]Samiul Alam , [1]Tahsin Reasat , [1]Asif Shahriyar Sushmit , [1]Sadi Mohammad Siddique
[3]Fuad Rahman , [4]Mahady Hasan , [1,*]Ahmed Imtiaz Humayun

[1]Bengali.AI, [2]Bangladesh University of Engineering and Technology,
[3]Apurba Technologies Inc., [4]Independent University, Bangladesh

## Abstract

*Latin has historically led the state-of-the-art in handwritten optical character recognition (OCR) research. Adapting existing systems from Latin to* alpha-syllabary *languages is particularly challenging due to a sharp contrast between their orthographies. The segmentation of graphical constituents corresponding to characters becomes significantly hard due to a cursive writing system and frequent use of diacritics in the alpha-syllabary family of languages. We propose a labeling scheme based on graphemes (linguistic segments of word formation) that makes segmentation inside alpha-syllabary words linear and present the first dataset of Bengali handwritten graphemes that are commonly used in everyday context. The dataset contains* 411*k curated samples of* 1295 *unique commonly used Bengali graphemes. Additionally, the test set contains* 900 *uncommon Bengali graphemes for out of dictionary performance evaluation. The dataset is open-sourced as a part of a public Handwritten Grapheme Classification Challenge on Kaggle to benchmark vision algorithms for multi-target grapheme classification. The unique graphemes present in this dataset are selected based on commonality in the Google Bengali ASR corpus. From competition proceedings, we see that deep learning methods can generalize to a large span of out of dictionary graphemes which are absent during training.*

## 1. Introduction

Speakers of languages from the alpha-syllabary or *Abugida* family comprise of up to 1.3 billion people across India, Bangladesh, and Thailand alone. There is significant academic and commercial interest in developing systems that can optically recognize handwritten text for such languages with numerous applications in e-commerce, security, digitization, and e-learning. In the alpha-syllabary writing system, each word is comprised of segments made of character units that are in phonemic sequence. These segments act as the smallest written unit in alpha-syllabary languages and are termed as *Graphemes* [14]; the term alpha-syllabary itself originates from the alphabet and syllabary qual-ities of graphemes [7]. Each grapheme comprises of a *grapheme root*, which can be one character or several characters combined as a conjunct. The term character is used interchangeably with unicode character throughout this text. Root characters may be accompanied by *vowel* or *consonant diacritics*- demarcations which correspond to phonemic extensions. To better understand the orthography, we can compare the English word *Proton* to its Bengali transliteration প্রোটন (Fig. 1). While in English the characters are horizontally arranged according to phonemic sequence, the first grapheme for both Bengali and Devanagari scripts have a sequence of glyphs that do not correspond to the linear arrangement of unicode characters or phonemes. As most OCR systems make a linear pass through a written line, we believe this non-linear positioning is important to consider when designing such systems for Bengali as well as other alpha-syllabary languages.

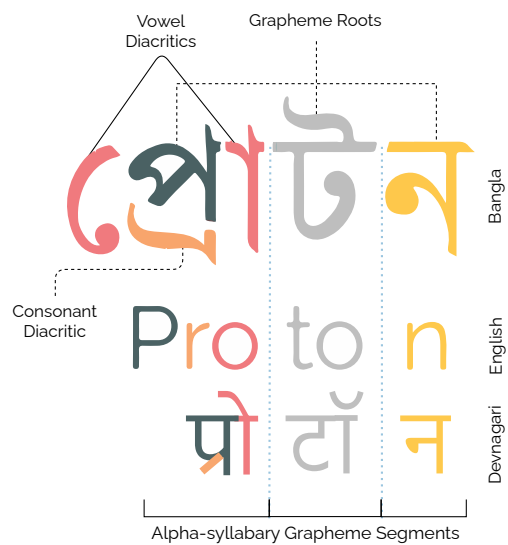

Figure 1: Orthographic components in a Bangla (Bengali) word compared to English and Devnagari (Hindi). The word 'Proton' in both Bengali and Hindi is equivalent to its transliteration. Characters are color-coded according to phonemic correspondence. Alpha-syllabary grapheme segments and corresponding characters from the three languages are segregated with the markers. While characters in English are arranged horizontally according to phonemic sequence, the order is not maintained in the other two languages.

We propose a labeling scheme based on grapheme segments of *Abugida* languages as a proxy for character based OCR systems; grapheme recognition instead of character recognition bypasses the complexities of character segmentation inside handwritten alpha-syllabary words. We have curated the first *Handwritten Grapheme Dataset* of Bengali as a candidate alpha-syllabary language, containing 411882 images of 1295 unique commonly used handwritten graphemes and ∼ 900 uncommon graphemes (exact numbers are excluded for the integrity of the test set). While the cardinality of graphemes is significantly larger than that of characters, through competition proceedings we show that the classification task is tractable even with a small number of graphemes- deep learning models can generalize to a large span of unique graphemes even if they are trained with a smaller set. Furthermore, the scope of this dataset is not limited to the domain of OCR, it also creates an opportunity to evaluate multi-target classification algorithms based on root and diacritic annotations of graphemes. Compared to Multi-Mnist [21] which is a frequently used synthetic dataset used for multi-target benchmarking, our dataset provides natural data of multi-target task comprising of three target variables.

The rest of the paper is organized as follows. Section 2 discuses previous works. Section 3 shows the different challenges that arise due to the orthography of the Bengali language, which is analogous to other *Abugida* languages. Section 4 formally defines the dataset objective, goes into the motivation behind a grapheme based labeling scheme, and discusses briefly the methodology followed to gather, extract, and standardize the data. Section 5 discusses some insights gained from the Bengali AI Grapheme Classification Competition along with solutions developed by the top-ranking participants. Finally section 6 presents our conclusions. For ease of reading, we have included the IPA standard English transliteration of Bengali characters in {.} throughout the text.

## 2. Related Work

**Handwritten OCR Datasets.** Although several datasets [2, 6, 20, 22] have been made for Bengali handwritten characters, their effectiveness has been limited. This could be attributed to the fact that they were formulated following contemporary datasets of English characters. All these datasets label individual characters or words. These datasets work well for English and can even be adapted for other document recognition tasks, setting the standard. Most character recognition datasets for other languages like the Devanagari Character Dataset [1, 19, 12] and the Arabic Printed Text Image Database [13, 3] were created following their design. However, they do not boast the same effectiveness and adaptability of their English counterparts. Languages with different writing systems therefore require language specific design of the recognition pipeline and need more understanding of how it affects performance. To the best of our knowledge, this is the first work that proposes grapheme level recognition for alpha-syllabary OCR.

## 3. Challenges of Bengali Orthography

As mentioned before in section 1, each Bengali word is comprised of segmental units called graphemes. Bengali has 48 characters in its alphabet- 11 vowels and 38 consonants (including special characters 'ৎ'{t̪},'ং' {ṁ},'ঃ'{h}). Out of the 11 vowels, 10 vowels have diacritic forms. There are also four consonant diacritics, 'ঘ' (from consonant ঘ {gha}), 'র' (from consonant র{ra}), '্র' (also from consonant র {ra}) and '়'. We follow the convention of considering 'ং'{ṁ},'ঃ' {h} as standalone consonants since they are always present at the end of a grapheme and can be considered a separate root character.

### 3.1. Grapheme Roots and Diacritics

Graphemes in Bengali consist of a root character which may be a vowel or a consonant or a consonant conjunct along with vowel and consonant diacritics whose occurrence is optional. These three symbols together make a grapheme in Bengali. The consonant and vowel diacritics can occur horizontally, vertically adjacent to the root or even surrounding the root (Fig. 2). These roots and diacritics cannot be identified in written text by parsing horizontally and detecting each glyph separately. Instead, one must look at the whole grapheme and identify them as separate targets. In light of this, our dataset labels individual graphemes with root characters, consonant and vowel diacritics as separate targets.

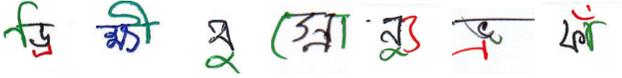

Figure 2: Different vowel diacritics (green) and consonant diacritics (red) used in Bengali orthography. The placement of the diacritics are not dependent on the grapheme root.

### 3.2. Consonant Conjuncts or Ligatures

Consonant conjuncts in Bengali are analogous to ligatures in Latin where multiple consonants combine together to form glyphs which may or may not contain characteristics from the standalone consonant glyphs. In Bengali, up to three consonants can combine to form consonant conjuncts. Consonant conjuncts may have two (second order conjuncts, eg. ষ্ট = শ + ট {sta = śa + ta }) or three (third order conjuncts, eg. ক্ষ্ণ = ক + ষ + ন {kṣṇa = ka + ṣa + na}) consonants in the cluster. Changes in the order of consonants in a conjunct may result in complete or partial changes in the glyph. The glyphs for conjuncts can get very complex and can even be hard for human subjects to discern (See Section C of Appendix in supplementary materials).

### 3.3. Allographs

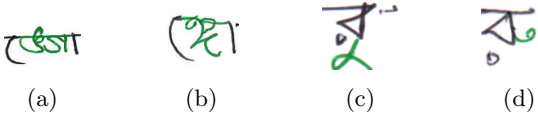

(a)  (b)  (c)  (d)

Figure 3: Examples of allograph pairs for the same consonant conjunct 'ঙ' {ṅga} (3a and 3b) and the same vowel diacritic 'ু' {u} (3c and 3d) marked in green. 3b and 3d follows an orthodox writing style.

It is also possible for the same grapheme to have multiple styles of writing, called allographs. Although they are indistinguishable both phonemically and in their unicode forms, allographs may in fact appear to be significantly different in their handwritten guise (Fig. 3). Allograph pairs are sometimes formed due to simplification or modernization of handwritten typographies, i.e. instead of using the orthodox form for the consonant conjunct ঙ = ঙ + গ as in Fig. 3b, a simplified more explicit form is written in Fig. 3a. The same can be seen for diacritics in Fig. 3c and Fig 3d. It can be argued that allographs portray the linguistic plasticity of handwritten Bengali.

### 3.4. Unique Grapheme Combinations

One challenge posed by grapheme recognition is the huge number of unique graphemes possible. Taking into account the 38 consonants ($n_c$) including three special characters, 11 vowels ($n_v$) and ($n_c^3 + n_c^2$) possible consonant conjuncts (considering $2^{nd}$ and $3^{rd}$ order), there can be $((n_c-3)^3 + (n_c-3)^2 + (n_c-3)) + 3$ different grapheme roots possible in Bengali. Grapheme roots can have any of the 10+1 vowel diacritics ($n_{vd}$) and 7+1 consonant diacritics ($n_{cd}$). So the approximate number of possible graphemes will be $n_v + 3 + ((n_c - 3)^3 + (n_c - 3)^2 + (n_c - 3)) \cdot n_{vd} \cdot n_{cd}$ or 3883894 unique graphemes. While this is a big number, not all of these combinations are viable or are used in practice.

## 4. The Dataset

Of all the possible graphemes combinations, only a small amount is prevalent in modern Bengali. In this section we discuss how we select the candidates for data collection and formalize the grapheme recognition problem.

### 4.1. Grapheme Selection

To find the popular graphemes, we use the text transcriptions for the Google Bengali ASR dataset [17] as our reference corpus. The ASR dataset contains a large volume of transcribed Bengali speech data. The transcription data by itself is very large and well standardized. It consists of 127565 utterances comprising 609510 words and 2111256 graphemes. Out of these graphemes, 1295 commonly used Bengali graphemes in everyday vocabulary is selected. Each candidate grapheme had to occur more than twice in the entire corpus or used in at least two unique words to be selected in our pool. Graphemes from highly frequent transliterations and misspelled words were also considered. The uncommon graphemes were synthesized by uniformly sampling from all the possible combinations and verifying their legibility. (See Section B of Appendix in supplementary materials for a full list of grapheme roots and diacritics)

### 4.2. Labeling Scheme

Bengali graphemes can have multiple characters depending on the number of consonants, vowels or diacritics forming the grapheme. We split the characters of a Bengali grapheme into three target variables based on their co-occurrence:

1. Vowel Diacritics, i.e. া, ি, ী, ু, ূ, ৃ, ে, ৈ, ো, ৌ. If the grapheme consists of a vowel diacritic, it is generally the final character in the unicode string. Graphemes cannot contain multiple vowel diacritics. The vowel diacritic target variable has 11 ($N_{vd}$) orthogonal classes including a null diacritic denoting absence.

2. Consonant Diacritics, i.e. ্য, ্র, ঁ, ্. Graphemes can have a combination of consonant diacritic characters eg. '্র্য' = '্র' + '্য'. We consider each combination to be a unique diacritic in our scheme for ease of analysis. The consonant diacritic target variable has 8 ($N_{cd}$) orthogonal classes including combinations and a null diacritic.

3. Grapheme roots, which can be comprised of vowels, consonants or conjuncts. In unicode these are placed as the first characters of a grapheme string. An alternative way of defining grapheme roots would be considering all the characters apart from diacritics as root characters in a grapheme. While possible orthogonal classes under this target variable can be a very big number (see Section 3.4), we limit the number of consonant conjuncts based on commonality in everyday context. There are in total 168 ($N_r$) roots in the dataset.

Grapheme recognition thus becomes a multi-target classification task for handwritten Bengali, rather than the traditional multi class classification adopted by previous datasets of this kind [20, 2, 22, 3]. Here, a vision algorithm would have to separately recognize grapheme roots, vowel diacritics and consonant diacritics as three target variables. Formally, we consider our dataset $\mathcal{D} = \{s_1, s_2, ..., s_N\}$ as composed of $N$ data points $s_i = \{x_i, y_i^r, y_i^{vd}, y_i^{cd}\}$. Each datum $s_i$ consists of an image, $x_i \in \mathbb{R}_{H \times W}$ and a subset of three target variables $y_i^r$, $y_i^{vd}$ and $y_i^{cd}$ denoting the ground truth for roots, vowel diacritics and consonant diacritics respectively, i.e., $y_i^r \in \mathcal{R}$, $y_i^{vd} \in \mathcal{V}$, and $y_i^{cd} \in \mathcal{C}$. Here, $\mathcal{R}$, $\mathcal{C}$, and $\mathcal{V}$ is the set of roots, vowel diacritics and consonant diacritics, respectively, where $|\mathcal{R}| = N_r$, $|\mathcal{C}| = N_{cd}$, and $|\mathcal{V}| = N_{vd}$. The multi-target classification task, thus will consist of generating a classifier $h$ which, given an image $x_i$, is capable of accurately predicting its corresponding components, i.e., $h(x_i) = \{y_i^r, y_i^{vd}, y_i^{cd}\}$.

Although we have formulated the grapheme recognition challenge as a multi-target classification task, it is only one way of defining the grapheme recognition problem. In fact, we will see in Section 5.2, that the problem can also be defined as a multi-label and a metric learning task.

### 4.3. Dataset collection & standardization

The data was obtained from Bengali speaking volunteers in schools, colleges and universities. A standardized form (See Section A of Appendix in supplementary materials) with alignment markers were printed and distributed. A total of 2896 volunteers participated in the project. Each subject could be uniquely identified through their institutional identification number submitted through the forms, which was later de-identified and replaced with a unique identifier for each subject. The dataset curation pipeline is illustrated in Fig. 4.

**Collection.** Contributors were given one of 16 different template forms with prompts for graphemes. The templates were automatically generated using Adobe Photoshop scripts. Each template had a unique set of graphemes compared to the others. Since the 16 different templates were not dispersed uniformly every time minor sampling bias was introduced during collection.

**Pruning and Scanning.** The forms were scanned and analysed carefully to remove invalid submissions and reduce sampling bias. In this step additional errors were introduced due to misalignment during scanning. Unfilled or improperly filled samples were still retained. All the forms were scanned using the same device at 300 dpi.

**Extraction.** An OCR algorithm was used to automatically detect the template ID. The template identifier asserted which ground truth graphemes where present in which boxes of the form. In this step, OCR extraction errors introduced label noise. The scanned forms were registered with digital templates to extract handwritten data, which sometimes introduced alignment errors or errors while extracting metadata. Unfilled boxes were removed automatically in this step.

**Preliminary Label Verification.** The extracted graphemes were compiled into batches and sent to 22 native Bengali volunteers who analysed each image and matched them to their corresponding ground truth annotation. In this step OCR errors and label noise was minimised. However additional error was introduced in the form of conformity bias, linguistic bias (i.e. allograph not recognized), grapheme bias (i.e. particular grapheme has significantly lesser number of samples) and annotator subjectivity. Samples selected as erroneous by each annotator was stored for further inspection instead of being discarded.

**Label Verification.** Each batch from the previous step, was sent to one of two curators who validated erroneous samples submitted by annotators and rechecked unique graphemes which had a higher fre-quency of mislabeled samples.

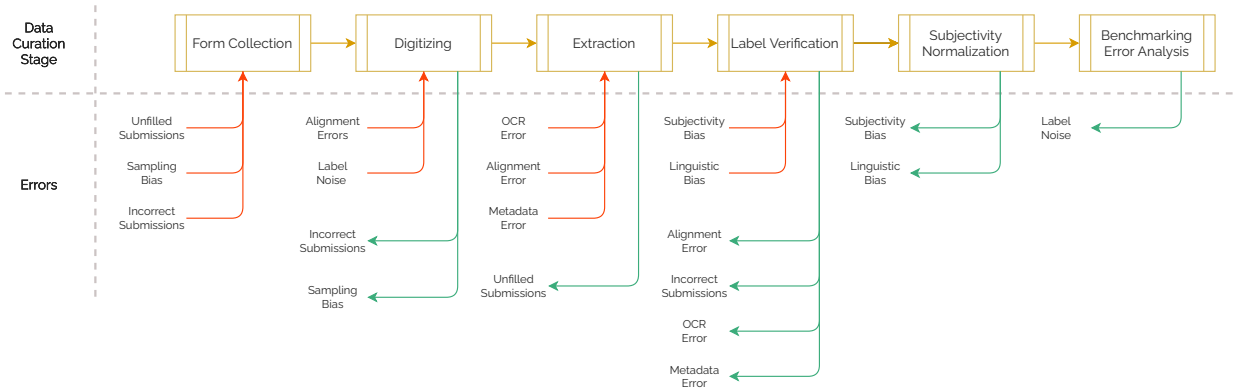

Figure 4: Overview of dataset creation process. Green arrows refer to the bias/errors removed in each step and red refers to the ones inevitably introduced.

Table 1: Number of samples present in each subset of the dataset. Null diacritics are ignored.

| Targets | Sub-targets | Classes | Samples | Training Set | Public Test Set | Private Test Set |
|---|---|---|---|---|---|---|
| Roots | Vowel Roots | 11 | 5315 | 2672 | 1270 | 1398 |
| | Consonant Roots | 38 | 215001 | 107103 | 52185 | 57787 |
| | Conjunct Roots | 119 | 184050 | 91065 | 45206 | 53196 |
| | **Total** | 168 | 404366 | 200840 | 98661 | 112381 |
| Diacritics | Vowel Diacritics | 10 | 320896 | 159332 | 78503 | 89891 |
| | Consonant Diacritics | 7 | 152010 | 75562 | 37301 | 44649 |

**Subjectivity Normalization.** A fixed guideline is decided upon by all curators that specifies how much and the nature of deformity a sample can contain. Based on this, subjectivity errors were minimized for unique graphemes with high frequency mislabeled samples.

### 4.4. Training Set Metadata

The metadata collected through forms are compiled together for further studies on dependency of handwriting with each of the meta domains. Only the training set metadata is made public; the test set metadata will be made available upon request. The training set contains handwriting from 1448 individuals, each individual contributing 138.8 graphemes on average; 1037 of the contributors identified as male, 383 as female, 4 as non-binary and 24 declined to identify. The medium of instruction during primary education for 1196 contributors was Bengali, for 214 English and for 12 Madrasha (Bengali and Arabic); 33 are left-handed while 1192 are right handed. Of all the contributors, 93 fall in the age group between $0 - 12$, 245 in $13 - 17$, 1057 in $18 - 24$, 22 in $25 - 35$ and 2 in ages between $36 - 50$.

### 4.5. Dataset Summary

A breakdown of the composition of the train and test sets of the dataset is given in Table 1. Additionally, a breakdown of the roots into vowels, consonants and conjuncts along with the number of unique classes and samples for each target is also shown. Note that the absence of a diacritic which is labeled as the null diacritic '0' is not considered when counting the total samples as the glyph for the diacritic is not present in such samples. The final dataset contains a total of 411882 handwritten graphemes of size 137 by 236 pixels. See supplementary materials and Appendix A for dataset collection forms, tools and protocols.

### 4.6. Class Imbalance in Dataset

We divide the roots into three groups- vowels, consonants, and consonant conjuncts- and inspect class imbalance within each. There are linguistic rules which constrict the number of diacritics that may occur with each of these roots, eg. vowel roots never have added diacritics. Although imbalance in vowel roots is not major, it must be noted because the relatively infrequent vowel roots 'ঈ', 'ঊ' $\{ī, ū\}$ and 'ঐ'$\{ai\}$ share a close resemblance to the more frequent roots 'ই', 'উ' $\{i,$

u} and 'এ' {ē} respectively.

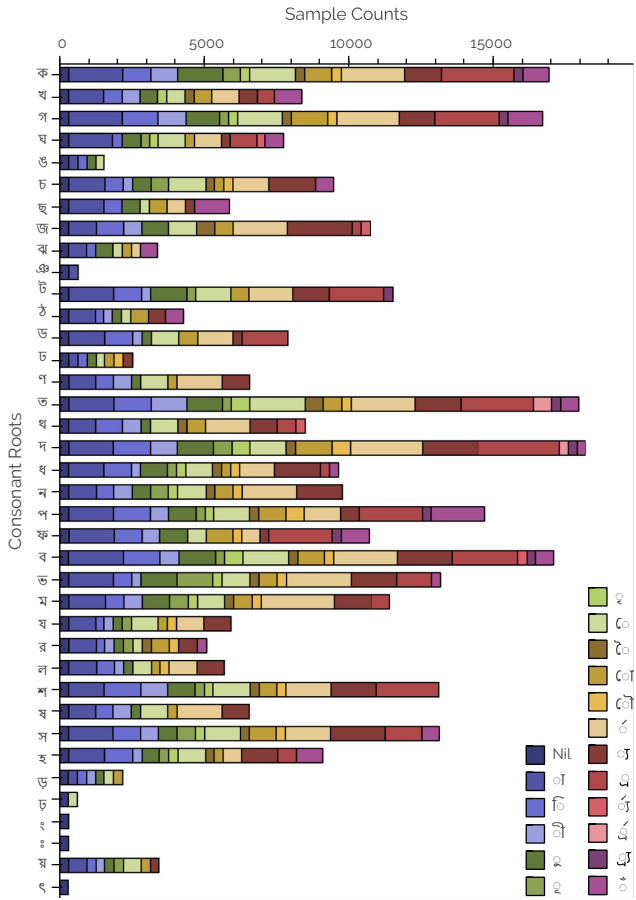

Figure 5: Number of samples per consonant root. Each bar represents the number of samples which contain a particular consonant and the divisions in each bar represent the distribution of diacritics in the samples containing that consonant.

The imbalance in consonant roots is however much more striking as we can see in Fig. 5. The imbalance here is twofold- in the number of total sample images of consonant roots and the imbalances in the distribution of the vowel and consonant diacritics that can occur with each consonant. The consonant conjuncts demonstrate imbalance similar to the consonant roots but with an added degree of complexity. We can visualize this imbalance much better via the chord diagram in Fig. 6. The consonant conjuncts are made up of multiple consonant characters and since the glyph of a consonant conjunct often shares some resemblance with its constituent consonants, highly frequent consonants may increase estimation bias for less frequent conjuncts containing them. This phenomenon is indeed visible and further discussed in Section. 5.3.

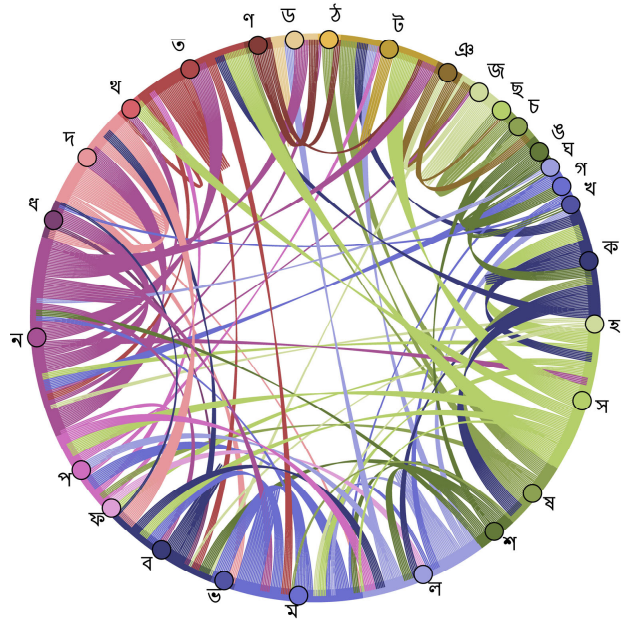

Figure 6: Connectivity graph between consonants forming second-order conjuncts. The length of each arc shows how often the consonant occurs as a consonant conjunct. Higher frequency of consonants in (eg. ক{ka}) may bias lower frequency conjuncts towards the constituent.

## 5. The Challenge

The dataset is open-sourced as part of a public Handwritten Grapheme Recognition Kaggle competition. Of all the samples present in the dataset, 200840 were placed in the training set, 98661 in the public test set, and 112381 in the private test set; while making sure there is no overlap in contributors between sets. Most of the uncommon graphemes were placed in the private test set and none in the training subset. Throughout the length of the competition, the participants try to improve their standings based on the public test set results. The private test set result on the other hand, is kept hidden for each submission and is only published after the competition is over. Of the OOD graphemes, 88.4% were placed in the private test set to prevent over-fitting models based on public standings. This motivated the participants to build methods that have the capacity to classify out of dictionary graphemes by recognizing the target variables independently.

### 5.1. Competition Metric

The metric for the challenge is a hierarchical macro-averaged recall. First, a standard macro-averaged re-call is calculated for each component. Let the macro-averaged recall for grapheme root, vowel diacritic, and consonant diacritic be denoted by $R_r$, $R_{vd}$, and $R_{cd}$ respectively. The final score $R$ is the weighted average

$$R = \frac{1}{4}(2R_r + R_{vd} + R_{cd}). \quad (1)$$

Table 2: Top 10 results on grapheme recognition challenge. Number of teams used separate models for in dictionary (ID) and out of dictionary (OOD) classification. MC = multi-class, ML = multi-label; MT = multi-target; SM = similarity metric learning.

| Rank | Augmentation | Problem Transform ID | Problem Transform OOD | OOD Detection | Model Architecture ID (M1) | Model Architecture OOD (M2) | Public Score | Private Score |
|---|---|---|---|---|---|---|---|---|
| 1 | Auto-Augment[9] | MC | MC | M1 confidence below threshold and M2 output is OOD | EfficientNet-b7[24] | CycleGan[27], EfficientNet-b7 | 0.995 | 0.976 |
| 2 | Fmix[15] | ML | - | - | SeResNeXt[16] | - | 0.996 | 0.969 |
| 3 | CutMix[25] | SM | MT | Arcface[10] | EfficientNet, SeResNeXt | EfficientNet | 0.995 | 0.965 |
| 4 | Cutout[11] | SM | MT | Arcface | Arcface | EfficientNet | 0.994 | 0.962 |
| 5 | CutMix | MT | - | - | SeResNeXt | - | 0.994 | 0.958 |
| 6 | CutMix, MixUp[26], Cutout, GridMask[8] | MT | - | - | SeResNeXt, InceptionResNetV2[23] | - | 0.987 | 0.956 |
| 7 | CutMix, Cutout | MT | - | - | PNASNet-5[18] | - | 0.994 | 0.955 |
| 8 | Cutout[11] | MT | MT | Arcface | EfficientNet-b7 | EfficientNet | 0.994 | 0.955 |
| 9 | CutMix, Grid-Mix[4] | MT | - | - | SeResNeXt, Arcface | - | 0.992 | 0.954 |
| 10 | Cutout | MT | - | - | SeResNeXt | - | 0.984 | 0.954 |

## 5.2. Top scoring methods

The Kaggle competition resulted in 31,002 submissions from 2,059 teams consisting of 2,623 competitors. The participants have explored a diverse set of algorithms throughout the competition(See Section E in Appendix for more details); the most popular being state of the art image augmentation methods such as cutout [11], mixup [26], cutmix [25], mixmatch [5] and fmix [15]. Ensemble methods incorporating snapshots of the same or different network architectures were also common.

The *winner* took a grapheme classification approach rather than component recognition. The input images were classified into 14784 (168 x 11 x 8) classes, that is, all the possibles graphemes that can be created using the available grapheme roots and diacritics. An EfficientNet [24] model is initially used to classify the graphemes. However, if the network is not confident about its prediction then the sample is considered as an OOD grapheme and it is passed on to the OOD grapheme classification pipeline. The pipeline consists of a CycleGan [27] that is trained to convert handwritten graphemes into typeface rendered graphemes. An EfficientNet classifier trained on a 14784 class synthetic grapheme dataset is used as an OOD grapheme classifier.

The *second* place team also built their solution upon a mulit-label grapheme classifier, but the number of classes were limited to the 1295 unique graphemes in training. A post processing heuristic is employed to generalize for any OOD graphemes. For each class in a target variable (eg. consonant diacritic 'ি'), probabilities of all the grapheme labels containing the target are averaged. This is repeated for every class for a target and the class with the highest average probability is selected for that target variable. Architectures used are different variants of SE-ResNeXt [16].

The *third* placed team dealt with the OOD graphemes by using metric learning. They used Arcface [10] to determine if an input test sample is present or absent in the train dataset. If the grapheme is present, a single EfficientNet model is used that detects the grapheme components in a multi-target setting. Otherwise, a different EfficientNet models are used to recognize each grapheme component.

A brief outline of the different approaches used by the participants and how they handled In Dictionary (ID) and Out of dictionary (OOD) graphemes are given in Table 2.

## 5.3. Submission Insights

For exploratory analysis on the competition submissions, we take the top 20 teams from both the private and public test set leaderboards and categorize their submissions according to quantile intervals of their private set scores as *Tier 1* ($> .976$), *Tier 2* ($> .933, < .976$), *Tier 3* ($> .925, < .933$) and *Tier 4* ($> .88, < .925$). It is seen that the *Tier 4* submissions have low discrepancy between public and private test set metrics; suggesting these to be high bias - low variance estimators. The *Tier 3* submissions were the total opposite and had high discrepancy on average, indicating fine-tuning of the model on the public test set. This discrepancy increases as we go to *Tier 2* submissions but then decreases for *Tier 1* submissions. In fact if we observe the progression of the *Tier 2* teams, many of them were once near the top of the private leaderboard but later fine-tuned for the public test set. Contrary to intuition, error rate doesn't significantly vary depending on the frequency of the unique grapheme in the training set, within each of these quantile groups. However, the error rate is higher by 1.31% on average for OOD graphemes, with *Tier 1* error rate at 4.4($\pm$0.07)%.

Considering the size of the challenge test set, possible reasons behind test set error could be label noise, class imbalance or general challenges surrounding the task. We start by inspecting the samples that were misclassified by all the *Tier 1* submissions and find that only 34.8% had label noise. Significant error can be seen due to the misclassification of consonant diacritic classes which are binary combinations of {'ো','ৗ' or 'ে'}, with high false positives for the individual constituents. This can be attributed to the class imbalance in the dataset since combinations are less frequent that their primitives; separately recognizing the primitives in a multi-target manner could be a possible way to reduce such error. The vowel diacritic component has the highest macro-averaged recall, proving to be the easiest among the three tasks.

The false negatives of different grapheme roots give us significant insights on the challenges present in Bengali orthography. A pair of grapheme roots can have high similarity due to common characters or even similarity between glyphs of constituents. Probing the *Tier 1* false negatives, we find that 56.5% of the error is between roots that share at least one character. Misclassification between consonant conjuncts with the same first and second characters accounts for 28.8% and 21.5% of the error. Confusion between roots by *Tier 1* submissions highly correlate with similarity between glyphs and is visualized in Fig. 7. Edges correspond to the sum of false negative rates between nodes. Edges are pruned if the sum is below .5%. Sub-networks are formed by groups that are similar to each other. Class imbalance also plays an interesting role in such cases; misclassification of roots with high similarity between their handwritten glyphs can also be biased towards one class due to higher frequency, eg. roots with 'ণ' {ṇa} are more frequently misclassified as roots with 'ন' {na} because of 'ন' {na} being more frequent. For more insight, see Section F in Appendix for confusion between roots and Section D.

## 6. Conclusion

In this paper, we outlined the challenges of recognizing Bengali handwritten text and explained why a character based labeling scheme- that has been widely successful for English characters- does not transfer well to Bengali. To rectify this, we propose a novel label-ing scheme based on graphemes and present a dataset based on this scheme. Crowd sourced benchmarking on Kaggle shows that algorithms trained on this dataset can generalize on out of dictionary graphemes. This proves that it is possible to summarize the entire cohort of graphemes through some representative samples. This grapheme labeling scheme could be used as a stepping stone to solve OCR related tasks in not only Bengali but also other related languages in the alpha-syllabary family.

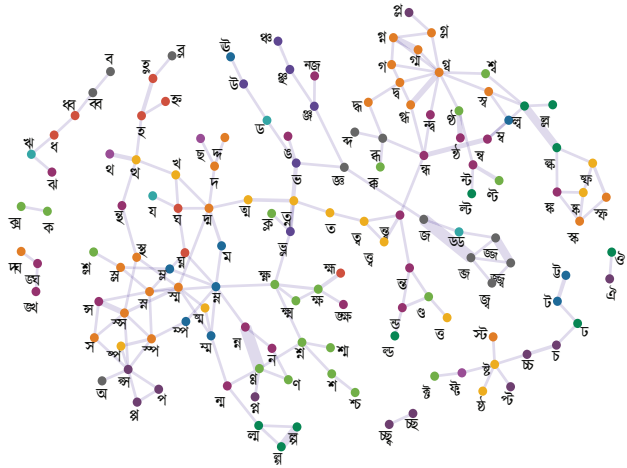

Figure 7: Similarity between handwritten grapheme roots based on *Tier 1* confusion. Nodes are color coded according to the first character in each root. Edges correspond to sum of false negative rates between nodes, higher edge width corresponds to higher similarity between handwritten grapheme roots.

## References

[1] S. Acharya, A. K. Pant, and P. K. Gyawali. Deep learning based large scale handwritten devanagari character recognition. In *Proc. 9th Int. Conf. SKIMA)*, pages 1–6, Dec 2015. 2

[2] Samiul Alam, Tahsin Reasat, Rashed Mohammad Doha, and Ahmed Imtiaz Humayun. Numtadb - assembled bengali handwritten digits. *arXiv preprint arXiv:1806.02452*, 2018. 2, 4

[3] Jawad H. AlKhateeb. A database for arabic handwritten character recognition. *Procedia Computer Science*, 65:556 – 561, 2015. 2, 4

[4] Kyungjune Baek, Duhyeon Bang, and Hyunjung Shim. Gridmix: Strong regularization through local context mapping. *Pattern Recognition*, 109:107594. 7

[5] David Berthelot, Nicholas Carlini, I. Goodfellow, Nicolas Papernot, A. Oliver, and Colin Raffel. Mixmatch: A holistic approach to semi-supervised learning. In *Proc. NeurIPS, 2019*, 2019. 7

[6] Mithun Biswas, Rafiqul Islam, Gautam Kumar Shom, Md Shopon, Nabeel Mohammed, Sifat Momen, and Md Anowarul Abedin. Banglalekha-isolated: A comprehensive bangla handwritten character dataset. *arXiv preprint arXiv:1703.10661*, 2017. 2

[7] William Bright. A matter of typology: Alphasyllabaries and abugidas. *Written Language & Literacy*, 2(1):45–65, 1999. 2

[8] Pengguang Chen. Gridmask data augmentation. *arXiv preprint arXiv:2001.04086*, 2020. 7

[9] Ekin D Cubuk, Barret Zoph, Dandelion Mane, Vijay Vasudevan, and Quoc V Le. Autoaugment: Learning augmentation policies from data. *arXiv preprint arXiv:1805.09501*, 2018. 7

[10] Jiankang Deng, Jia Guo, Niannan Xue, and Stefanos Zafeiriou. Arcface: Additive angular margin loss for deep face recognition. In *Proc. 2019 IEEE/CVF Conf. CVPR*, June 2019. 7

[11] Terrance DeVries and Graham W Taylor. Improved regularization of convolutional neural networks with cutout. *arXiv preprint arXiv:1708.04552*, 2017. 7

[12] K. Dutta, P. Krishnan, M. Mathew, and C. V. Jawahar. Offline handwriting recognition on devanagari using a new benchmark dataset. In *Proc. 13th IAPR Int. Workshop DAS*, pages 25–30, April 2018. 2

[13] Reza Farrahi Moghaddam, Mohamed Cheriet, Mathias M. Adankon, Kostyantyn Filonenko, and Robert Wisnovsky. Ibn sina: A database for research on processing and understanding of arabic manuscripts images. In *Proc. 9th IAPR Int. Workshop DAS*, page 11–18, 2010. 2

[14] Liudmila Fedorova. The development of graphic representation in abugida writing: The akshara's grammar. *Lingua Posnaniensis*, 55(2):49–66, Dec 2013. 1

[15] Ethan Harris, Antonia Marcu, Matthew Painter, Mahesan Niranjan, Adam Prügel-Bennett, and Jonathon Hare. FMix: Enhancing Mixed Sample Data Augmentation. *arXiv e-prints*, page arXiv:2002.12047, Feb. 2020. 7

[16] J. Hu, L. Shen, and G. Sun. Squeeze-and-excitation networks. In *Proc. IEEE/CVF Conf. CVPR, 2018*, pages 7132–7141, June 2018. 7

[17] Oddur Kjartansson, Supheakmungkol Sarin, Knot Pipatsrisawat, Martin Jansche, and Linne Ha. Crowdsourced speech corpora for javanese, sundanese, sinhala, nepali, and bangladeshi bengali. In *Proc. 6th Intl. Workshop SLTU*, pages 52–55, 08 2018. 3

[18] Chenxi Liu, Barret Zoph, Maxim Neumann, Jonathon Shlens, Wei Hua, Li-Jia Li, Li Fei-Fei, Alan Yuille, Jonathan Huang, and Kevin Murphy. Progressive neural architecture search. In *Proceedings of the European Conference on Computer Vision (ECCV)*, pages 19–34, 2018. 7

[19] Sk Md Obaidullah, Chayan Halder, Nibaran Das, and Kaushik Roy. A new dataset of word-level offline handwritten numeral images from four official indic scripts and its benchmarking using image transform fusion. *Int. J. Intell. Eng. Inform.*, 4(1):1–20, Feb. 2016. 2

[20] AKM Shahariar Azad Rabby, Sadeka Haque, Md. Sanzidul Islam, Sheikh Abujar, and Syed Akhter Hossain. Ekush: A multipurpose and multitype comprehensive database for online off-line bangla handwritten characters. In *RTIP2R*, pages 149–158, 2019. 2, 4

[21] Sara Sabour, Nicholas Frosst, and Geoffrey E Hinton. Dynamic routing between capsules. In *Advances in neural information processing systems*, pages 3856–3866, 2017. 2

[22] Ram Sarkar, Nibaran Das, Subhadip Basu, Mahantapas Kundu, Mita Nasipuri, and Dipak Kumar Basu. Cmaterdb1: a database of unconstrained handwritten bangla and bangla–english mixed script document image. *IJDAR*, 15(1):71–83, 2012. 2, 4

[23] Christian Szegedy, Sergey Ioffe, Vincent Vanhoucke, and Alex Alemi. Inception-v4, inception-resnet and the impact of residual connections on learning. *arXiv preprint arXiv:1602.07261*, 2016. 7

[24] Mingxing Tan and Quoc Le. EfficientNet: Rethinking model scaling for convolutional neural networks. In *PMLR*, volume 97, pages 6105–6114, 09–15 Jun 2019. 7

[25] S. Yun, D. Han, S. Chun, S. J. Oh, Y. Yoo, and J. Choe. Cutmix: Regularization strategy to train strong classifiers with localizable features. In *Proc. ICCV, 2019*, pages 6022–6031, Oct 2019. 7

[26] Hongyi Zhang, Moustapha Cisse, Yann N Dauphin, and David Lopez-Paz. Mixup: Beyond empirical risk minimization. *arXiv preprint arXiv:1710.09412*, 2017. 7

[27] J. Zhu, T. Park, P. Isola, and A. A. Efros. Unpaired image-to-image translation using cycle-consistent adversarial networks. In *Proc. ICCV, 2017*, pages 2242–2251, Oct 2017. 7

# Appendix

## A. Graheme Collection Form

Handwritten Graphemes were collected from 16 different template forms designed for efficient extraction by scanning. Forms also had additional information that include chirality, age group, gender, medium of instruction and location of primary school to allow for further study into handwriting. Names and IDs were collected to keep into account if multiple forms were submitted by the same person. All forms and data were later de-identified. A sample form is showed in Fig. 8

Figure 8: Sample of OCR form for extracting handwritten graphemes

## B. Label-Class Overview in Dataset

All classes for each of the three target variables have been listed in their utf-8 form in Table 4. Note that all the diacritics are shown in a grapheme with root ব as an example. Their glyphs will, however, appear unchanged with any other root.

## C. Contribution Error and Subjectivity

The original data went through a rigorous curation process; approximately 12817 samples were discarded due to either label noise or corrupted submissions. Samples that would be illegible to human annotators without prior knowledge of the ground truth were discarded from the dataset. A GUI based python toolkit was provided for the label verification process (see Fig. 12). Although this was done to make sure the data is clean, it should be mentioned that a concrete definition of which samples should be considered legible does not exist. In fact some would consider a written sample perfectly legible while others would consider the same as absolutely unclear. If we look at the error data, among the top 100 unique graphemes that had the most erroneous contributions, 83 had consonant conjuncts as the grapheme root. Six out of the nine 3$^{rd}$ order consonant conjuncts in the dataset, were among the most erroneous graphemes. This matches the intuition that more complex grapheme glyphs are harder to discern as writers are more likely to make mistakes in typography when writing glyphs with more intricate patterns. The most common errors in writing graphemes categorized by the number of simple consonants in the root are given in Table 3.

Table 3: Frequently discarded graphemes during label verification phases of dataset curation.

| Root Category | Top 5 Mislabeled Graphemes | | | | |
|---|---|---|---|---|---|
| Consonants | গুঁ | গুঁ | গু | শ্যু | ঘু |
| 2nd order Conjuncts | ষ্ট্যা | ক্কৃ | স্পা | ক্ষ | ক্রী |
| 3rd order Conjuncts | ঞ্জিত্র | ক্ষ্ম | স্ক্রৌ | জ্ঞা | ন্দ্বি |

## D. Consonant Conjuncts vs Diacritics

One question that arises while splitting the constituents of a grapheme into the three bins of our labeling scheme (see Section 4.2) is the ambiguity between consonant conjuncts and consonant diacritics. While conjuncts are formed by adding multiple consonants together, consonant diacritics also add consonants with other consonants but as demarcations that are completely different from the original glyph of the consonant. For example, the consonant diacritic 'ৢ' is completely different from its original form 'য'. Whenever it is added to a consonant root, the root retains its original glyph. This is not always the case for consonant conjuncts, where the consonants being added to might change its form significantly. In Bengali grammar, consonant diacritics are called *Fola* and defined separately from consonant conjuncts as *Jukto Borno*. The consonants that have diacritic forms do not construct conjuncts, eg. 'য' and 'র' are not present as a second order conjunct constituent in Fig. 6.

Table 4: Table of all target variables and classes present in the dataset. Diacritics are written in their graphemic forms with grapheme root 'ব'. Phonetic transliterations in IPA standards are provided in brackets (·).

| Target Variable | Class | |
|---|---|---|
| Grapheme roots (168) | VOWEL ROOTS | |
| | অ (a), আ (ā), ই (i), ঈ (ī), উ (u), ঊ (ū), ঋ (r), এ (ē), ঐ (ai), ও (ō), ঔ (au) | |
| | CONSONANT ROOTS | |
| | ক (ka), খ (kha), গ (ga), ঘ (gha), ঙ (ṅa), চ (ca), ছ (cha), জ (ja), ঝ (jha), ঞ (ña), ট (ṭa), ঠ (ṭha), ড (ḍa), ঢ (ḍha), ণ (ṇa), ত (ta), থ (tha), দ (da), ধ (dha), ন (na), প (pa), ফ (pha), ব (ba), ভ (bha), ম (ma), য (ya), র (ra), ল (la), শ (śa), ষ (ṣa), স (sa), হ (ha), ড় (ṛa), ঢ় (ṛha), য় (ẏa), ◌ং (ṁ), ◌ঃ (ḥ), ৎ (ṯ) | |
| | CONJUNCT ROOTS | |
| | ক্ক (kka), ক্ট (kṭa), ক্ত (kta), ক্ল (kla), ক্ষ (kṣa), ক্ষ্ণ (kṣṇa), ক্ষ্ম (kṣma), ক্স (ksa), গ্ধ (gdha), গ্ন (gna), গ্ব (gba), গ্ম (gma), গ্ল (gla), ঘ্ন (ghna), ঙ্ক (ṅka), ঙ্ক্ত (ṅkta), ঙ্ক্ষ (ṅkṣa), ঙ্খ (ṅkha), ঙ্গ (ṅga), ঙ্ঘ (ṅgha), চ্চ (cca), চ্ছ (ccha), চ্ছ্ব (cchba), জ্জ (jja), জ্জ্ব (jjba), জ্ঞ (jña), জ্ব (jba), ঞ্চ (ñca), ঞ্ছ (ñcha), ঞ্জ (ñja), ট্ট (ṭṭa), ড্ড (ḍḍa), ণ্ট (ṇṭa), ণ্ঠ (ṇṭha), ণ্ড (ṇḍa), ণ্ণ (ṇṇa), ত্ত (tta), ত্ত্ব (ttba), ত্থ (t'tha), ত্ন (tna), ত্ব (tba), ত্ম (tma), দ্ঘ (dgha), দ্দ (dda), দ্ধ (d'dha), দ্ব (dba), দ্ভ (dbha), দ্ম (dma), ধ্ব (dhba), ন্জ (nja), ন্ট (nṭa), ন্ঠ (nṭha), ন্ড (nḍa), ন্ত (nta), ন্ত্ব (ntba), ন্থ (ntha), ন্দ (nda), ন্দ্ব (ndba), ন্ধ (ndha), ন্ন (nna), ন্ব (nba), ন্ম (nma), ন্স (nsa), প্ট (pṭa), প্ত (pta), প্ন (pna), প্প (ppa), প্ল (pla), প্স (psa), ফ্ট (phṭa), ফ্ফ (phpha), ফ্ল (phla), ব্জ (bja), ব্দ (bda), ব্ধ (bdha), ব্ব (bba), ব্ল (bla), ভ্ল (bhla), ম্ন (mna), ম্প (mpa), ম্ব (mba), ম্ভ (mbha), ম্ম (m'ma), ম্ল (mla), ল্ক (lka), ল্গ (lga), ল্ট (lṭa), ল্ড (lḍa), ল্প (lpa), ল্ব (lba), ল্ম (lma), ল্ল (lla), শ্চ (śca), শ্ন (śna), শ্ব (śba), শ্ম (śma), শ্ল (śla), ষ্ক (ṣka), ষ্ট (ṣṭa), ষ্ঠ (ṣṭha), ষ্ণ (ṣṇa), ষ্প (ṣpa), ষ্ফ (ṣpha), ষ্ম (ṣma), স্ক (ska), স্ট (sṭa), স্ত (sta), স্থ (stha), স্ন (sna), স্প (spa), স্ফ (spha), স্ব (sba), স্ম (sma), স্ল (sla), স্স (s'sa), হ্ন (hna), হ্ব (hba), হ্ম (hma), হ্ল (hla) | |
| Vowel Diacritics (11) | Null, বা (bā), বি (bi), বী (bī), বু (bu), বূ (bū), বে (bē), বৈ (bai), বো (bō), বৌ (bau) | |
| Consonant Diacritics (8) | Null, ব্য (bya), ব্র (bra), র্ব (rba), র্ব্য (rbya), ব্র্য (brya), র্ব্র (rbra), বঁ (b̃) | |

## E. Rank Progression in Competition

To benchmark the dataset extensively, a Kaggle competition was organized based on this dataset. This resulted in 31,002 submissions from 2,059 teams consisting of 2,623 competitors. Analyzing the submissions allowed us to further discover important statistics on this dataset. Since the competition was held with half the test set hidden, it is important to know when the model is over-fitting and find ways to generalize for OOD data. Without proper tuning, large models will tend to over-fit on the public test set and perform poorly on the private one. This is shown in Fig. 10. Five teams' progression has been highlighted and the others grayed out. Notice how the teams marked ■ and ■ were able to decrease their score discrepancy and jump up in rank. The teams marked ■, ■ and ■, scored high in the private test set at some point but over-fit thereafter falling in rank.

A major reason for many of the teams over-fitting to the public set is that the test set distribution is somewhat different from the training distribution. The problem is compounded by the presence of OOD graphemes in the test set. These account for about 3.5% of test samples. So it is crucial that steps be taken to check if models trained on the dataset use augmentation and generative techniques to match the distribution of the test set. The issue of generalization can be further checked by seeing how a model performs on samples that are comparatively less fre-quent in the training set. We took all submissions that were reasonably good (top 75% among all submissions corresponding to a private score of 0.88 and higher) and classified them further into four classes - *Tier 1* (Top 25%, score > 0.976), *Tier 2* (Between 25% and 50%, $0.933 < \text{score} < 0.976$), *Tier 3* ((Between 50% and 75%, $0.925 < \text{score} < 0.933$) and *Tier 4* (Bottom 25%, $0.88 < \text{score} < 0.925$). We mapped their performance against their public-private score discrepancy defined as the difference, $\Delta$ between their public and private scores as shown in Table 5. The discrepancy levels were categorized according to their quantile intervals and the number of submissions belonging to that interval are shown in the table. Notice how the *Tier 4* submissions category have low discrepancy. These submissions were high bias - low variance. The *Tier 3* submissions category were the total opposite and had mostly high discrepancy. These submissions were over-fitting to the training data. However, notice the transition that occurs between *Tier 2* and the *Tier 1* models in terms of the change in the number of low and high discrepancy submissions. The *Tier 2* submissions were from models that were over optimized for the public test set which degraded their performance on the private set.

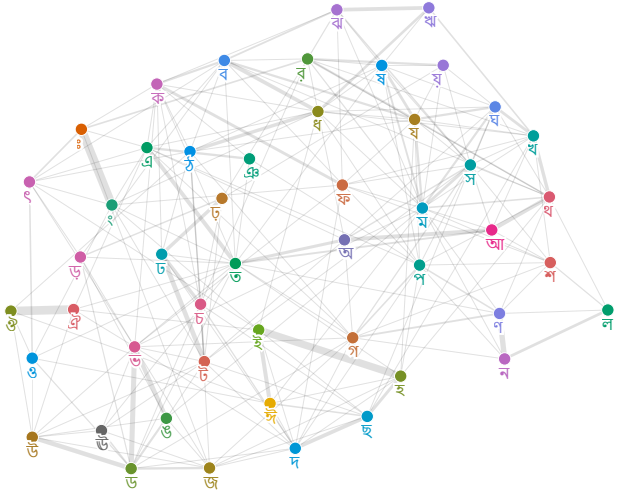

Figure 9: Similarity between handwritten non-conjunct roots based on Tier 1 confusion scores. Edges correspond to sum of false negative rates between nodes, higher edge width corresponds to higher similarity. Only roots with an error rate more than 15% are shown in the figure.

Table 5: Public-Private score discrepancy in the submissions by Top 20 teams

| Performance Tiers | Public-Private Discrepancy, $\Delta$ | Number of Submissions |
|---|---|---|
| Tier 4 | $\Delta \leq 0.04$ | 392 |
|  | $0.4 < \Delta < 0.06$ | 211 |
|  | $\Delta \geq 0.06$ | 116 |
| Tier 3 | $\Delta \leq 0.04$ | 100 |
|  | $0.04 < \Delta < 0.06$ | 158 |
|  | $\Delta \geq 0.06$ | 461 |
| Tier 2 | $\Delta \leq 0.04$ | 27 |
|  | $0.04 < \Delta < 0.06$ | 191 |
|  | $\Delta \geq 0.06$ | 500 |
| Tier 1 | $\Delta \leq 0.04$ | 201 |
|  | $0.04 < \Delta < 0.06$ | 159 |
|  | $\Delta \geq 0.06$ | 361 |

Another important aspect to see is how the different models were performing with regard to the number of samples present in the training set. We categorized the frequency of each unique graphemes in the training set, $f$ according to quartile division. The the specific intervals are $f > 165$ (the top 25% most frequent graphemes in the training set) $157 < f \leq 165$, $143 < f \leq 157$, $0 < f \leq 143$(the bottom 25%). Additionally, the graphemes not present in training set were correspond to $f = 0$. The performance of each Tier is tabulated in Table 6. One thing to notice here is that the error rate of graphemes is significantly higher for the *Tier 4* class (The error rate here is the percentage of graphemes the submission predicted incorrectly and is not the same as the competition score). Additionally, better submissions perform better in general. However, the error rate on the OOD set is still higher than the other categories for all of the Tiers.

Table 6: Generalization performance with respect to different grapheme training frequencies. f=0 corresponds to OOD graphemes.

| Performance Tiers | Training frequency, $f$ | Test Error Rate(%) |
|---|---|---|
| Tier 4 | $f > 165$ | 28.42 |
|  | $157 < f \leq 165$ | 28.34 |
|  | $143 < f \leq 157$ | 28.42 |
|  | $0 < f \leq 143$ | 28.42 |
|  | $f = 0$ | 29.68 |
| Tier 3 | $f > 165$ | 4.47 |
|  | $157 < f \leq 165$ | 4.36 |
|  | $143 < f \leq 157$ | 4.48 |
|  | $0 < f \leq 143$ | 4.48 |
|  | $f = 0$ | 6.02 |
| Tier 2 | $f > 165$ | 3.86 |
|  | $157 < f \leq 165$ | 3.77 |
|  | $143 < f \leq 157$ | 3.90 |
|  | $0 < f \leq 143$ | 3.87 |
|  | $f = 0$ | 5.27 |
| Tier 1 | $f > 165$ | 3.37 |
|  | $157 < f \leq 165$ | 3.29 |
|  | $143 < f \leq 157$ | 3.41 |
|  | $0 < f \leq 143$ | 3.41 |
|  | $f = 0$ | 4.44 |

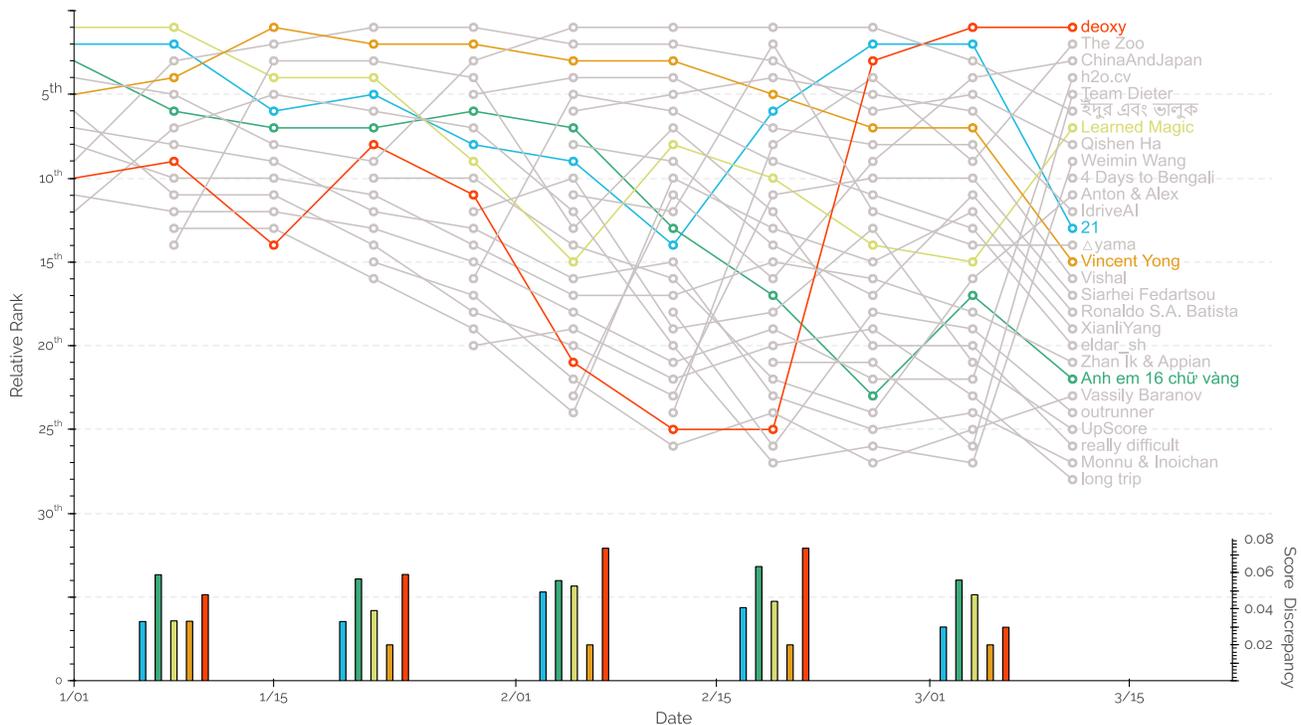

Figure 10: Figure showing how the relative rank of selected teams changed during the competition. Some teams have been highlighted and their corresponding public-private score difference/discrepancy represented by the bar diagrams.

## F. Confusion between consonant and vowel roots of Tier 1 submissions

It has been suggested in Section 4.6 that class imbalance could cause models to focus on features of roots and diacritics that are more frequent in the dataset. Due to nature of Bengali orthography, many glyphs share features that are very similar. This causes model to confuse an infrequent grapheme in the dataset to be misclassified as a more frequent one. It is therefore essential to understand which features of a glyph the models are putting importance on. This may be approximated by looking at how much each consonant root has been misclassified in the form of a confusion network graph shown in Fig. 9. We can notice here that many of the misclassified roots are expected. For example, হ and ই are miscalssified more frequently by the *Tier 1* models and this can be attributed to the similarity of their glyphs. However, there are other consonant roots like ঔ and ঐ which do not share strong similarity but are still misclassified frequently. This suggests that the models could also focus on specific parts of the characters for similarity instead of their primary structure. Fig. 9 also suggest that in handwritten form, a lot of these consonants have reduced difference in their glyphs compared to their printed form. This can also be used as a guideline for future datasets as well as rule based methods to take into account the natural similarity between handwritten graphemes.

## G. Confusion between Conjunct roots with common consonants

The presence of common consonants between conjunct roots influence the confusion between them. If we look at the the error rate of graphemes with consonant conjuncts as roots, we will see that a majority of the graphemes are predicted as a different conjunct with a common character. Of the erroneous predictions made by Tier 1 models, 56.1% of the error is due to confusion between conjunct roots that have a common character; 28..8% of the total error is due to confusion between roots with the same first character. This can be clearly seen in Figure 11, which shows the false negatives by all the Tier 1 submissions. The false negatives can be seen to form a block diagonal arrangement in Figure 11a, when the roots are ordered alphabetically. This arrangement results in conjuncts with the same first character like 'দ্দ' 'দ্দ' 'দ্ব' and 'দ্ব' to become adjacent. The similarity between adjacent roots with the same first character can be seen in 11b and 11c. High false negative off diagonal blocks can correspond to either confusion due to the same second character in a root or confusion due to similar looking first characters (Fig. 9). Example of such a case can be seen in Fig. 11e for similar looking first characters ন and ণ.

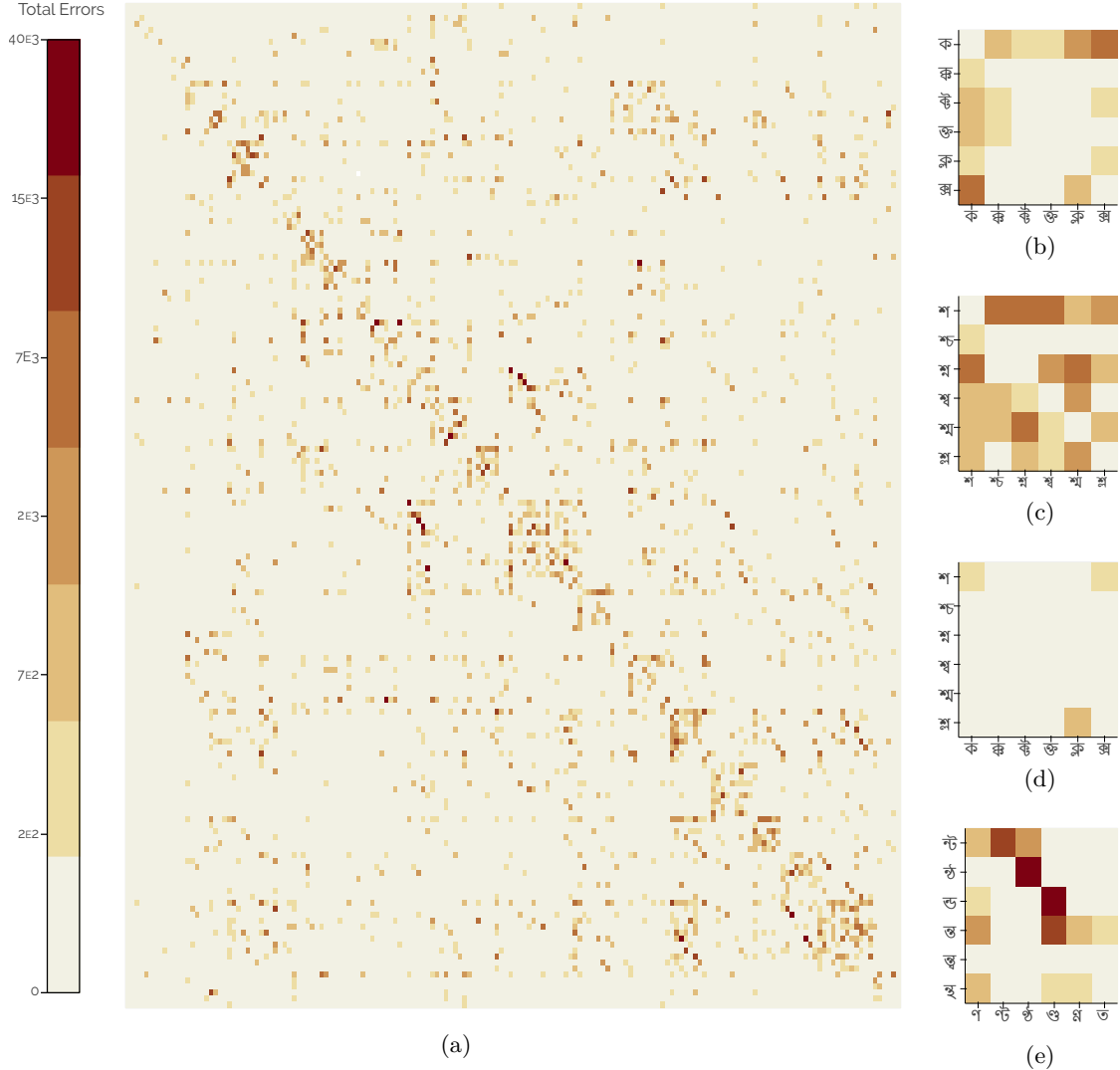

Figure 11: False Negatives Matrix for roots compiled from all 664 *Tier*1 submissions. The roots are arranged in aphabetic order. a shows the entire matrix. Notice the significant number of errors near the diagonal indicating that the first character's glyph is more prominent in corresponding conjuncts. Shown here, b and c are blocks running along the diagonal with roots 'ক','ক্র','ক্ট','ক্ত','ক্ল','ক্ষ' and 'শ','শ্চ','শ্ম','শ্ব','শ্য','শ্ল' respectively. d shows a sample off diagonal block of roots 'ক','ক্র','ক্ট','ক্ত','ক্ল','ক্ষ' incorrectly predicted as 'শ','শ্চ','শ্ম','শ্ব','শ্য','শ্ল'. Here the number of false negatives are very low compared to the diagonal blocks as the glyphs are not very similar with the exception of শ্ল and ক্ল due to same second character. However, there are cases where off diagonal blocks show large errors due to stark similarity in glyphs as shown in e.

Figure 12: Label verification tool for handwritten graphemes (see 4.3). The tool was used by curators to check individual samples for errors. Samples with similar ground truth are arranged sequentially along columns; to make sure that this does not induce conformity bias, annotators had the option to hide the ground truth labels (top-left corner in each grapheme block). Marked samples using this tool are moved to a separate error folder after annotation is complete. Annotators were also encouraged to revisit their error folders after each annotation pass to normalize subjective choices between early and latter samples.